%% file: arxiv-pgmanalysis-2018.tex
\documentclass{article}

\pdfoutput=1

\usepackage{microtype}
\usepackage{booktabs} 

\usepackage[utf8]{inputenc} 
\usepackage[T1]{fontenc}    
\usepackage{url}            
\usepackage{amsfonts}       
\usepackage{nicefrac}       
\usepackage[pdftex]{graphicx}
\usepackage{wrapfig}
\usepackage{caption}
\usepackage{enumitem}
\usepackage{amsmath}
\usepackage{array}

\usepackage{subfigure,subfloat}
\usepackage{listings}
\usepackage{color}
\usepackage[english]{babel}
\usepackage{multirow}
\usepackage{mdframed}

\usepackage{textcomp}
\usepackage[none]{hyphenat}

\lstdefinestyle{C++}{
language=[ISO]{C++},
basicstyle=\ttfamily\scriptsize
}

\usepackage{hyperref}

\usepackage[accepted]{ml2019}

\mltitlerunning{Using Structured Input and Modularity for Improved Learning}
\begin{document}

\twocolumn[
\mltitle{Using Structured Input and Modularity for Improved Learning}



\mlsetsymbol{equal}{*}

\begin{mlauthorlist}
\mlauthor{Zehra Sura}{to}
\mlauthor{Tong Chen}{to}
\mlauthor{Hyojin Sung}{to}
\end{mlauthorlist}

\mlaffiliation{to}{IBM Research, Yorktown Heights, NY, USA}

\mlcorrespondingauthor{Zehra Sura}{zehrasura@gmail.com}

\mlkeywords{program analysis, deep learning, neural networks}

\vskip 0.3in

\begin{abstract}
  We describe a method for 
  utilizing the known structure of input data 
  to make learning more efficient.
  Our work is in the domain of programming languages,
  and we use deep neural networks to do program analysis.
  Computer programs include a lot of structural information 
  (such as loop nests, conditional blocks, and data scopes),
  which is pertinent to program analysis.
  In this case, the neural network has to learn to recognize the structure, 
  and also learn the target function for the problem.
  However, the structural information in this domain is 
  readily accessible to software with the availability of 
  compiler tools and parsers for well-defined programming languages.
  
  Our method for 
  utilizing the known structure of input data 
  includes: (1) pre-processing the input data 
  to expose relevant structures, and (2) constructing neural networks
  by incorporating the structure of the input data 
  as an integral part of the network design. 
  The method has the effect of modularizing the neural network 
  which helps break down complexity, 
  and results in more efficient training of the overall network. 
  We apply this method to an example code analysis problem, 
  and show that it can achieve higher accuracy with a smaller network size 
  and fewer training examples.
  Further, the method is robust, 
  performing equally well on input data with different distributions.
\end{abstract}
]



\printAffiliationsAndNotice{}

\section{Introduction}
\label{sec:intro}
\input{intro}

\section{Our Approach}
\label{sec:design}
\input{design}

\section{Modular Design Method}
\label{sec:modconst}
\input{modconst}

\section{Experiments}
\label{sec:exp}
\input{exp}

\section{Related Work}
\label{sec:related}
\input{related}

\section{Conclusion}
\label{sec:conc}
\input{conc}

\nocite{*}

\bibliography{arxiv-pgmanalysis-2018}
\bibliographystyle{ml2019}

\end{document}

%% file: intro.tex
In this paper we focus on using deep learning for computer program analysis and
code optimization, which are essential for performance.
Achieving high performance for an application program can be complex, 
made even more complex by the intricate and heterogeneous hardware platforms in use today. 
Software tools such as compilers and runtime systems are used to manage this complexity. 
These tools rely on accurate program analysis to optimize code and deliver performance. 
Some of the analysis and optimization decisions are simple with 
clear and efficient computation models,
but some are complex or even unknown (i.e.~no analytical model exists). 
In the latter case, heuristic models are used in the computation. 
The heuristics need to be developed using knowledge of the hardware characteristics and application domain, 
but it can be costly or infeasible to develop good heuristics because of   
the intractable number of factors and system interactions 
that impact performance.
Instead, the heuristics can be substituted with machine learning techniques to incorporate a data-driven approach in program analysis and optimization.

There is a lot of prior work on applying machine learning techniques 
in program analysis and optimization, 
for example choosing the strategies and parameters for loop optimization and parallelization, making decisions on code placement and scheduling, and choosing the set and ordering of transformations to apply.
In \cite{2018LitReview}, the authors present a comprehensive 
literature review on using machine learning for parallel systems  
covering work published since the year 2000.
For each relevant paper in their survey, the authors identify 
the objective, machine learning techniques applied, 
and code/system features considered as part of the training input data.
In \cite{2018SurveyPaper}, the authors survey the progress of work in the 
area of machine learning for compiler optimization.

Much of this prior work uses a list of static and/or dynamic program features
as input data, or else it directly uses the program code in the format of 
a string of characters.
We take a different approach where we apply domain knowledge to 
determine structures in the code (e.g.~loops, basic blocks, statements, etc.) 
that are relevant for the machine learning. 
We encode this structural information in the input data, 
which is now represented as a sequence of tokens.
For this purpose, we re-use functionality in compiler toolchains 
that are already good at extracting structural information in programs. 
By encoding the structural information explicitly in the training data, 
we make it easier for the automated machine learning to learn from it.
The data representation we use has several advantages:
\begin{enumerate}
    \item It provides input data directly in the form of code structures rather than lists of secondary features derived from the code.
    \item It provides expressibility: the encoded structures can be derived from the program code using custom rules and they need not directly match existing structures defined in the programming language.
    \item It allows flexibility in encoding additional property information associated with different structures, and allows embedding them at appropriate contextual points in the input data sequence.
\end{enumerate}

We are interested in exploring whether machine learning can be applied to 
more complex problems in program analysis and code optimization, 
therefore we base our work on using deep learning 
which is known to be a powerful machine learning technique.
For example, consider the problem of estimating the 
dynamic instruction count of a program.
The instruction count may be different for different types of operations, 
e.g.~a scalar arithmetic operation, a vector computation, or a memory access operation.
When the input program is a sequence of operations with no control flow, 
only localized properties in the sequence (i.e.~type of operation) contribute 
to the instruction count, and current deep learning methods perform very well in this case.
Likewise, when the input program has very simple control flow with only one level nested loops, the information needed to compute the instruction count is relatively easy to find in the program sequence (iteration count of the closest enclosing loop), and 
current LSTM-based deep learning again performs very well.
However, when the structure of the program gets more complicated, 
it becomes more difficult to extract the relevant contextual information 
from a program sequence
(e.g.~iteration counts of all enclosing loops). 
In this case, the performance of current LSTM-based techniques falls short and  accuracy of the learned models is low.
To address this, we designed a new technique for using 
deep neural networks (which we call \emph{modular neural networks}) 
that is capable of processing complicated program structures 
encoded in the input data to train more efficiently.

The main idea in our work is to translate high-level knowledge 
of the problem domain  
into the design (layout) of the deep neural network.
Each input item in the training data is a sequence of tokens, 
where a token corresponds to 
a high-level semantic structure in the problem domain. 
For example, in the case of program analysis, 
the types of structures may include a type corresponding to 
the beginning of a loop, the end of a loop, or a basic block of statements.
A modular neural network is composed of 
multiple component neural networks,
with one component neural network corresponding to each structure type.
The modular neural network includes control logic to 
route individual tokens within an input sequence 
to the appropriate component neural network.
The control logic also incorporates interconnection rules 
for updating the neural network state 
before/after the processing of each individual token. 
These interconnection rules are determined using knowledge of the problem domain.

Our modular design approach addresses several of the challenges highlighted by the study in \cite{2018LitReview}, 
including the following challenges that are important for program analysis:
\begin{enumerate}
\item Complexity: Prior work has successfully demonstrated the use of machine learning for some individual problems in isolation (e.g.~deciding whether to use a GPU or not for select kernels\cite{Fonseca2013}). However, the ability to solve more complicated problems using machine learning is still unknown (e.g.~using a sequence of transformations for overall optimization of loop nests).
In general, more powerful learning is needed to tackle complex problems.
Our modular approach is specifically designed to break down the complexity of learning.
\item Training data: Often the size of training data is limited for problems in the area of program analysis. There have been some efforts to collect and maintain common data repositories for this purpose\cite{CollectiveKnowledge, LoopRepo}, but these efforts are still nascent. 
Our modular approach is able to learn faster, achieving high accuracy 
using less training data.
\item Generality: Many prior works can only handle specific programming models or domain specific languages, or else they rely on programmer annotations.
Our technique is applicable to all programming languages.
\end{enumerate}

In this paper we make the following contributions:
\begin{itemize}
    \item We introduce a novel method of designing neural networks using modular neural networks. Our method naturally incorporates the relevant  information about program structures (domain knowledge) into the design of the neural network itself. The resulting network is capable of handling more complex learning problems than a monolithic network of equivalent (or larger) size designed using current techniques.
    \item We compare 
    an LSTM-based deep neural network designed 
    using current state-of-the-art techniques 
    with 
    a modular neural network designed using our method
    to solve an example code analysis problem. We show that our method is able to achieve higher accuracy with less training data and with an overall smaller size of network.
\end{itemize}

In Section~\ref{sec:design}, 
we describe how relevant structural information in computer programs 
can be made explicit in the input data representation 
and can be used to design a modular neural network.
In Section~\ref{sec:modconst}, we give a systematic method for 
our technique to use structural information and modularity in deep learning.
In Section~\ref{sec:exp} we detail our experiments and results.
In Section~\ref{sec:related} we describe related work and 
in Section~\ref{sec:conc} we present our conclusions.

%% file: design.tex
There are two main considerations when designing a neural network based solution: (1) the input data and the representation used for it, and (2) the design of the neural network structure. 
In current practice, 
the choices made in the design of these two aspects 
are relatively independent.
However, in our design approach the two are tightly integrated 
and the choice of input data representation directly impacts the design of the neural network as well.
In this section, we highlight the advantages of our method for both of these aspects of the design.

\subsection{Input Data Representation}
\label{sec:back:datarep}
When applying machine learning for program analysis and optimization, 
typically the input data is a set of features that capture the 
characteristics of the program code and/or the system platform being used for program execution. The features can be either static or dynamic.
For example, static code features may include number of basic blocks, 
number of branch instructions, etc.;
dynamic code features may include loop iteration counts, branch frequencies, 
total number of instructions, etc.;
static system features may include cache sizes, peak flops, etc.; dynamic system features may include values of various hardware counters.
The literature review in \cite{2018LitReview} gives a detailed listing of the 
various input features used in prior work.
The problem with using a set of such features as input is that these 
features are secondary properties inferred by the designer to be 
relevant to the objective of the machine learning.
However, they may not completely capture the first-order properties that 
directly impact the learning objective.
For example, if the objective is to predict the execution time of 
programs on a hardware device, the first-order input property 
is the program code, and the first-order output property is the observed runtime for that code when executed on the hardware device.

More recent work such as \cite{8091247} directly uses the program code 
as training input by encoding the program as a 
sequence of tokens embedded in vector space.
The problem with this approach is that the training input 
is represented at a very high-level, 
and it requires powerful machine learning capability to redact 
the parts of the input that are not relevant to the learning objective. 
Prior work has applied this technique to relatively simple problems 
and it is unclear if it will work for complex problems.

In our approach, the input is a sequence of tokens representing program structures. 
The specific program structures to be encoded into the sequence 
are selectively chosen depending on the objective function to be learned.
Figure \ref{fig:exampledatarep} shows an example to illustrate 
how input data can be represented.
The tokens for the identified program structures are encoded using integers, 
and they can optionally be followed by another value.
In the example, the \emph{BB} and \emph{LB} tokens take integer values and the \emph{IF} token takes a float value, as illustrated in the figure.
These optional values make it easy to embed specific contextual information 
at appropriate points in the program sequence.
Also, even though the example uses structures that map directly to 
traditional programming language constructs,
in general the identified structures can be more complex 
(e.g.~instances of a LoopBegin immediately followed by an If statement 
can be treated as a single structure), 
and can be tailored to be specific to the function being learned.
Existing functionality in compiler toolchains,
specifically lexers and parsers 
that translate programs to an intermediate format, 
can be used to process program code 
and generate the required input sequences.

\begin{figure}[ht!]
\small
\begin{mdframed}
  {\bf Objective Function}: \\
  \hspace*{0.5cm} Estimate dynamic instruction count\\
  {\bf Structures in Input}: \\
  \hspace*{0.5cm} BasicBlock({\em BB}), LoopBegin({\em LB}), LoopEnd({\em LE}), \\
  \hspace*{0.5cm} If({\em IF}), ElseIf({\em ELS}), EndIf({\em EIF})\\
  {\bf Example Code}:
  \begin{lstlisting}[frame=none]%[numbers=none]
  
  int x = rand(); 
  for (int i=0; i< N; i++) { 
    int y = rand();
    if (y > M)
        x += y;
  }
  return x;
  \end{lstlisting}
  {\bf Input Data Sequence}:\\
  \hspace*{0.5cm} {\em BB} $n_1$ {\em LB} $n_2$ {\em BB} $n_3$ {\em IF} $n_4$ {\em BB} $n_5$ {\em EIF} {\em LE} {\em BB} $n_6$ \\
  \hspace*{0.5cm} $n_1$,$n_5$,$n_6$ = instructions to compute line 1,5,7\\
  \hspace*{0.5cm} $n_2$ = iteration count for loop on line 2\\
  \hspace*{0.5cm} $n_3$ = instructions to compute lines 3 and 4\\
  \hspace*{0.5cm} $n_4$ = branch taken probability for line 4\\
  {\bf Expected Output}:\\
  \hspace*{0.5cm} $n_1 + n_2 * (n_3 + n_4*n_5) + n_6$
\end{mdframed}
\caption{Example to illustrate our input data representation}
\label{fig:exampledatarep}
\end{figure}

For the simple example in Figure~\ref{fig:exampledatarep}, 
the objective function is known and can be easily computed 
as part of an analytical program analysis model.
However, the machine learning approach uses data to \emph{learn} this function, 
and the same method can be applied to cases where the objective function 
is unknown, for example when the objective is to estimate the 
number of execution cycles 
for a processor supporting speculative out-of-order execution 
instead of the dynamic instruction count.

For our input data representation,
a designer with domain-specific knowledge 
can think in terms of the 
first-order properties of the problem being addressed, 
and need not derive a list of secondary properties.
At the same time, the input data is now curated to be 
more specific to the objective being learned, 
and this helps reduce the complexity of the problem for automated learning.
Thus, our approach balances 
the requirements on input data collection and processing 
with the limitations of automated machine learning. 
In general, choosing the correct level of balance between data processing 
and automated learning is an active research topic.

\subsection{Neural Network Design}
\label{sec:back:nndesign}
Network design determines the structure of the neural network which includes the type, number, and organization of the network layers and the interconnections between layers.
There has been much progress in training and using 
sophisticated deep learning networks, 
including the use of recursive neural networks or RNNs\cite{Jain:1999:RNN:553011} 
and Long Short Term Memory networks or LSTMs\cite{Hochreiter:1997:LSTM}.
These networks have been successfully applied to 
problem domains such as natural language processing(NLP) and machine translation \cite{10.1007/978-3-642-40585-3_14, nmt}.
However, it is still unknown how well they apply to 
complex problems in the domain of program analysis.

\begin{figure}[t]
\begin{center}
\includegraphics[width=0.75\linewidth]{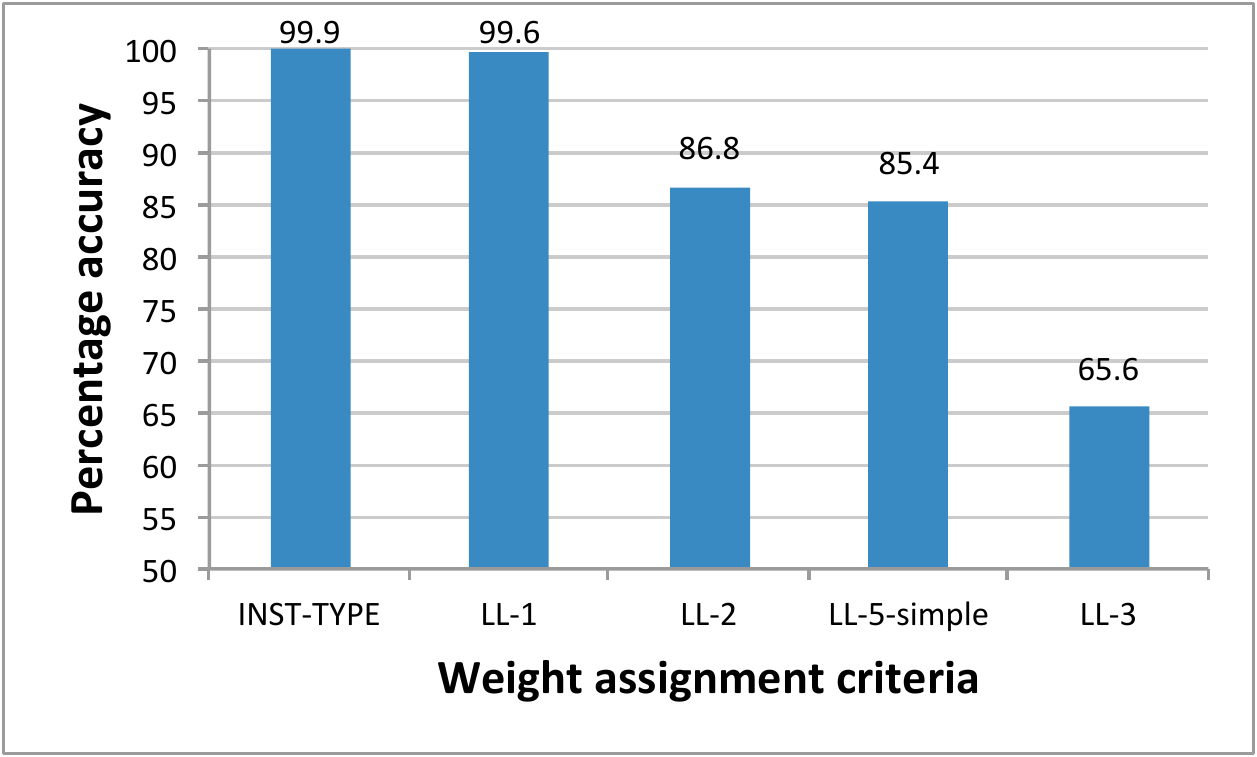}
\caption{Effect of objective function complexity on accuracy}
\label{fig:motivation}
\end{center}
\vspace{-0.2in}
\end{figure}

In our work, we first did some experiments to 
assess the capability of current well known deep learning techniques in our application context.
For our experiments, we chose the objective function to be 
a weighted count of instructions in the program, 
where the weights are determined based on some programming constructs.
We used different criteria for assigning weights, 
including the following:
\begin{enumerate}
    \item {\bf INST-TYPE}: weight is based on the type of instruction (e.g.~arithmetic or memory operation).
    \item {\bf LL-1}: weight is based on the iteration counts of enclosing loops, with only 1-level loop nests in the dataset.
    \item {\bf LL-2}: weight is based on the iteration counts of enclosing loops, with upto 2-level loop nests in the dataset.
    \item {\bf LL-3}: weight is based on the iteration counts of enclosing loops, with upto 3-level loop nests in the dataset.
    \item {\bf LL-5-simple}: weight is based on the nesting depth of the closest enclosing loop, with upto 5-level loop nests in the dataset.
\end{enumerate}
We trained neural network models, one for each of these criteria.
We carefully tuned the network designs to achieve maximum accuracy, 
using current state-of-the-art techniques 
and a network structure composed of a number of LSTM layers 
followed by a number of fully connected (FC) layers.
Our data generation and experimental methodologies are described in 
Sections~\ref{sec:datagen} and~\ref{sec:expmeth}.
Figure~\ref{fig:motivation} shows the  
accuracies obtained for the different trained models.

We observe that for the simpler criteria, the accuracy achieved is 
close to 100\%, but as the complexity of the problem increases, 
the accuracy drops off.
This motivated us to develop new methods to help improve the 
automated learning process.
We aim for a neural network design that has 
better learning capability (higher accuracy with less training data), 
so that we can apply it to progressively more complex problems.

Our approach is to use modularity to break down the complexity 
of learning.
As an example, consider the problem of computing weighted instruction counts 
using the nesting depth of the enclosing loop as the weight.
We can modularize the problem by breaking it down into three sub-computation parts: (1) computation on a LoopBegin (i.e.~increment a current nesting depth counter), 
(2) computation on a LoopEnd (i.e.~decrement the current nesting depth counter), and 
(3) computation on an instruction or BasicBlock of instructions (i.e.~accumulate a running count of instructions appropriately).
For an automated machine learning process, 
the computations for all three parts are unknown and have to be learned, 
but the three parts are distinct and it is easier to learn them individually than altogether as a monolithic function.
Note that the three parts correspond to distinct structures in the program, 
and our input data representation is tailored to explicitly identify 
these structures.
Our neural network design 
has some layers dedicated to learning each of the sub-computations separately.
The next section describes in detail a general method for constructing such modular networks.

%% file: modconst.tex
The black-box approach to neural network learning 
is to train the neural network with a sufficient amount of data, 
and let the network automatically figure out 
what the relevant information in the data is.
Instead of using a purely data-driven black-box approach, 
a second more discriminate approach is to pre-process the input data 
according to the learning objective 
(e.g. by annotating the data, or by using different encoding techniques), 
in order to make the relevant information explicit in the input data
(\citet{Kolokolov:2002:SPS:591227.591286,7981748}).
We introduce a third approach where some relevant information is 
encoded in the neural network itself. 
In our approach, the neural network is designed such that the 
network structure reflects the structural information in the input data domain.

Our approach is applicable when:
\begin{enumerate}[itemsep=1pt, topsep=1pt, parsep=1pt, leftmargin=18pt]
\item the input data domain is structured, i.e.~there exists a known formal grammar 
that defines all possible instances of the input data, and 
\item the learning objective has a dependency on the structure of the input data.
\end{enumerate}
Both of these conditions are satisfied for many problems in the area of program analysis, with the programming language grammar readily available.

We define a method for constructing {\em modular neural networks}, 
which are networks designed to utilize 
the structure within individual instances of input data. 
The objective is to learn a function $F(x)$, where $x$ is a member of grammar $G$. 
Our method for constructing a modular neural network is a four-step process 
described in the following subsections.

\subsection{Step 1: Identify Relevant Structures}
Since the learning objective depends on structure 
within individual instances of the input data, 
we first apply domain knowledge to identify a set of relevant structures ($S$). 
These structures will be explicitly marked in the input data.
Knowing the problem domain and the application requirements 
allows choosing structures at the right level of abstraction.
The structures are expressed using tokens and expressions from the known grammar $G$.
A grammar $G'$ can be derived based on $G$ such that there is one 
token in $G'$ for each member of $S$.
Then, for each input data $x$ member of $G$ there is a corresponding $x'$ member of $G'$, 
and $x'$ is a sequence of token values $t_{ij}$,
where $i$ denotes the position of $t$ in the sequence
and $j$ denotes the structure $s_{j}$ in set $S$ that the token $t$ corresponds to.

\subsection{Step 2: Pre-process Data}
We pre-process each input data instance $x$ (member of $G$) to transform it 
to the corresponding $x'$ (member of $G'$).
This transformation process can be easily automated
using existing tools (e.g.~compiler toolchains).
For each structure $s_{j}$ in set $S$, 
it can optionally be defined to carry extra property information, 
e.g.~LoopBegin structures may be defined to carry an integer value 
giving the loop iteration count.
The format of this extra information can be tailored individually for each structure $s_{j}$ in set $S$,
without worrying about normalizing the range or using a uniform size across all structures.
This is because the design of our neural network is modular 
and it uses separate component neural networks to process different structures in the input data.
Thus $x'$ is a sequence of token values $t_{ij}$, 
where for specific values of j, the token $t_{ij}$ is followed by an annotation value $v_{ij}$.
The objective now is transformed to learn the function $F(x')$.

\subsection{Step 3: Associate Sub-functions}
We associate a function $F_{j}$ 
with each $s_{j}$ that is a member of $S$.
We define a new hyperparameter in the overall neural network called {\em state vector} or $SV$.
The state vector is used to keep track of and to pass along 
historical contextual information between modular components of the neural network.
Each $F_{j}$ takes as input a current state vector value, and 
produces one output which is used to compute the next state vector value.
If there is an annotation value attached to instances of $s_{j}$, then 
the corresponding $F_{j}$ will take the annotation value as a second input.
In the common case, there is one neural network corresponding to each $s_{j}$.
However, it is possible to have flexible mappings from $s_{j}$ to component neural networks, 
including 1-to-1 and many-to-1.
It is also possible to use known predefined functions for some $F_{j}$, 
while learning only the remaining subset of functions.
This flexibility makes it possible to 
utilize domain knowledge 
and improve the design of the neural network 
to make learning more efficient.

\subsection{Step 4: Define Composition Rules}
For each $F_{j}$, we define a composition rule $F'_{j}$. 
The composition rules primarily influence how the input state vector for $F_{j}$ 
is determined, and how the output of $F_{j}$ is used 
to compute the new value of the state vector ($SV$).
As an example, the composition rules can define simple sequential semantics 
where the input state vector is the current value of $SV$, and the output is the 
next value of $SV$, i.e.~$F'_{j}(x)$ is mapped to ${SV = F_{j}(SV, x)}$.
In the common case, the same composition rules apply uniformly 
to all functions $F_{j}$ that are to be learned.
However, the rules can be tailored for specific functions based on semantics 
of the application domain.
For example, in the case of 
functions that operate on recursive structures,
the rules can include a stack mechanism to help track 
the flow of information across nested constructs. 
Consider $s_{1}=LoopBegin$ and $s_{2}=LoopEnd$. 
Then 
$F'_{1}(x)$ can be mapped to $\{SV = F_{1}(SV, x);~Stack\_Push(SV)\}$, 
and $F'_{2}(x)$ can be mapped to $\{SV = Stack\_Pop();~SV = F_{2}(SV, x)\}$. 

The initial value of the state vector $SV$ is a hyperparameter of the overall neural network.
The final constructed neural network has 
the individual neural networks 
for each $F_{j}$ as its components.

For a given input $x'=t_{1j_{1}}t_{2j_{2}}...t_{nj_{n}}$, 
this neural network learns (or infers):\\
$F(x')=F'_{j_{n}}(t_{nj_{n}}$, $F'_{j_{(n-1)}}(t_{(n-1)j_{(n-1)}}, ... F'_{j_{1}}(t_{1j_{1}}, \_ )...))$.
In comparison, a traditional RNN  works on input $x=t_{1}t_{2}...t_{n}$ and learns (or infers):\\ 
$F(x)=F(t_{n}, F(t_{n-1}, ... F(t_{1}, \_ ) ... ))$.

For backward propagation during training, the deltas are computed and propagated in 
reverse through the individual components that compose the overall neural network.
For this to work properly, the following conditions must be met:
\begin{enumerate}[itemsep=1pt, topsep=1pt, parsep=1pt, leftmargin=18pt]
\item Neural networks for functions to be learned must use differentiable components, as in current practice.
\item Functions that are predefined must either have an inverse, or a corresponding 
reverse relationship must be defined for all points in the data domain.
\item Composition rules must have a corresponding reverse rule, e.g.~$Stack\_Push$ in the 
forward pass transforms to a $Stack\_Pop$ in the backward pass, and vice versa.
\end{enumerate}

\subsection{Network Design}
\begin{figure}[t!]
\begin{center}
\includegraphics[width=0.85\linewidth]{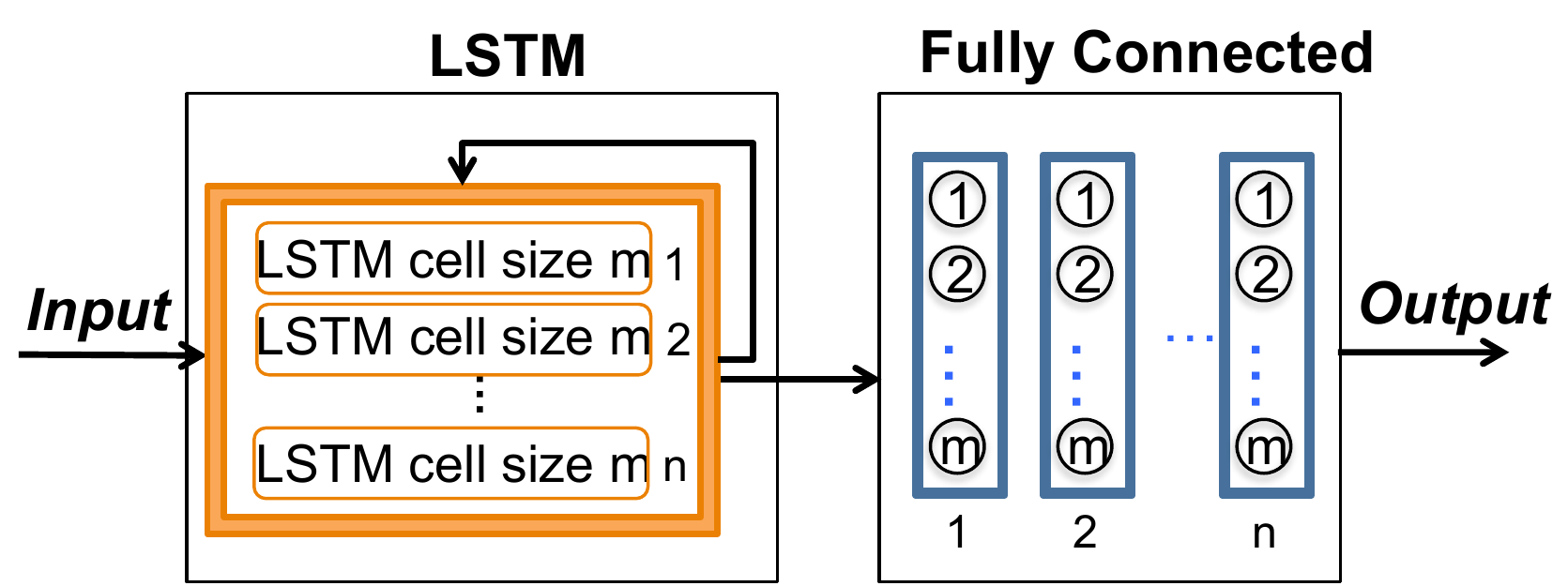}
\caption{Baseline network design}
\label{fig:tradnndesign}
\end{center}
\end{figure}

Our input data is a sequence of tokens, 
so we used LSTMs in our network design.
The network structure for the baseline comparison 
is composed of a number of LSTM layers, 
followed by a number of fully connected (FC) layers, 
as shown in Figure~\ref{fig:tradnndesign}.

\begin{figure}[t!]
\begin{center}
\includegraphics[width=0.99\linewidth]{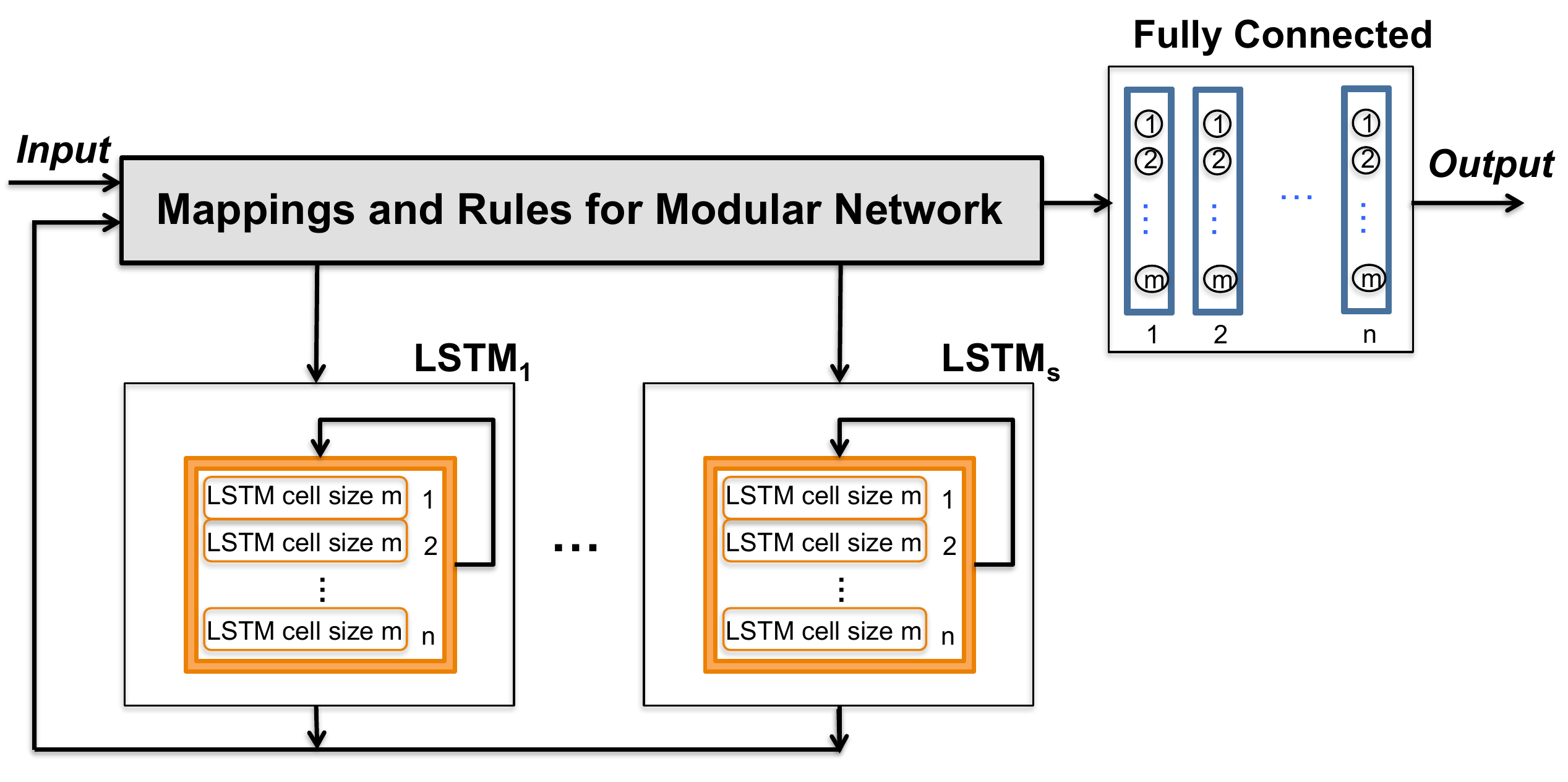}
\caption{Modular network design}
\label{fig:modnndesign}
\end{center}
\vspace{-0.2in}
\end{figure}

Figure~\ref{fig:modnndesign} illustrates the network design 
for our modular method. 
There is an instance of the LSTM network for each 
function $F_{j}$ to be learned.
The fully connected (FC) layers are placed at the end 
for output generation,
analogous to the baseline network design.
The {\em Mappings and Rules for Modular Network} 
represents the logic in functions $F'_{j}$
that governs how inputs and outputs are routed between 
the component LSTMs as the input data sequence is processed.

%% file: exp.tex
The goal of our experiments is to compare the learning capability of 
the three approaches described in the beginning of Section~\ref{sec:modconst}: 
(i) purely data-driven black-box neural networks, 
(ii) neural networks with discriminate data pre-processing, and 
(iii) modular neural networks with network structures designed in conjunction with the format of the input data. 
We perform a controlled experiment using a simple objective function and  
try to minimize the impact from factors such as feature selection, 
collection of training examples, and choice of neural network design parameters.
We first describe the learning objective, then describe our experimental methodology, 
and finally present our results.

\subsection{Learning Objective}
\label{sec:lprob}
\input{lprob}

\subsection{Methodology}
\label{sec:expmeth}
\input{expmeth}

\subsection{Results}
\label{sec:results}
\input{results}

%% file: lprob.tex
To perform a detailed analysis of the learning capability of our modular method 
we choose a problem that is complex enough to challenge the baseline method, 
is relevant to real problems in program analysis, 
and yet is simple enough to allow us to perform a controlled set of experiments.
We assume a programming language that 
includes loops and statements among other structured constructs.
In the case of loop nests, 
we observed from Figure~\ref{fig:motivation} that there are two factors contributing to complexity: 
(1) the non-linearity introduced by multiplying iteration counts of multiple loops (as shown by the drop in accuracy for LL-2 and LL-3), 
and (2) the structural variations of having loop nests of varying lengths (as shown by the drop in accuracy for LL-5-simple).
For the experiments in this paper, we focus on the second factor i.e.~complexity due to structure.
We define the function to be learned as a weighted count of instructions in the program, 
where the weight for each instruction is 
the loop nesting depth of that instruction.
This function has analog output values.
In our experiments, when computing the loss function or when testing for accuracy,
we characterize the output value to be correct if it is within 
a relative error range of the exact numerical value 
(set to 5\% in experiments).

The training data comprises of a set of computer programs in 
a programming language that is defined by grammar $G$.
These programs can be parsed using an existing compiler toolchain 
and converted into a sequence of tokens in $G$.
We identify three structures in the set $S$: $LoopBegin$ (LB), $LoopEnd$ (LE), and $BasicBlock$ (BB).
$LoopBegin$ and $LoopEnd$ are also expressions/tokens in grammar $G$, 
whereas $BasicBlock$ is only defined in the derived grammar $G'$, 
which is as follows:
\begin{align*} 
&Program~\rightarrow~StmtList \\
&StmtList~\rightarrow~Stmts~|~StmtList~Stmts \\
&Stmts~\rightarrow~\textbf{BasicBlock}~|~Loop \\
&Loop~\rightarrow~\textbf{LoopBegin}~StmtList~\textbf{LoopEnd} 
\end{align*}
All three structures in $S$ are tokens in grammar $G'$.
The transformed input data ($x'$ in $G'$) contains 
sequences of tokens corresponding to the three identified structures, 
with each token followed by a property value.
The property value for $BasicBlock$ is the number of instructions in the basic block.
The property value for $LoopBegin$ and $LoopEnd$ is the loop iteration count.

We compare four different methods for using neural networks to learn the objective function:

\noindent{\bf Baseline }
This is the black-box method that relies on the neural network to automatically 
figure out the relevant information from the input data.

\noindent{\bf Modular }
This is our newly introduced method that identifies relevant structures in the data, 
and constructs a modular neural network based on the identified structures.
For our experimental evaluation, both the {\em Baseline} and {\em Modular} methods use 
an identical input data format.

\noindent{\bf PartiallyExplicit }
This method pre-processes the data to explicitly mark the outermost boundary 
of loop nests.
It introduces 2 new tokens: $LoopNestBegin$ (LNB) and $LoopNestEnd$ (LNE). 
The input data instances are transformed to insert these new tokens 
before the $LoopBegin$ and $LoopEnd$ tokens corresponding only to the outermost loops.
As an example, the following sequence (shown without the annotation values):\\
\hspace*{0.5cm} BB\ \ LB\ \ LB\ \ BB\ \ LE\ \ BB\ \ LE\ \ LB\ \ BB\ \ LE\\
is tranformed to:\\
\hspace*{0.5cm} BB\ \ $LNB$\ \ LB\ \ LB\ \ BB\ \ LE\ \ BB\ \ $LNE$\ \ LE\ \ $LNB$\ \ LB\ \ BB\ \ $LNE$\ \ LE\\
The input data instances used in this method have a longer 
sequence length than the input data instances used in all other methods,
with redundant information added into the input data.
The data pre-processing introduced in this approach is somewhat generic, 
in that it can apply to multiple learning objectives in the domain of program analysis.

\noindent{\bf FullyExplicit}
This method pre-processes the data to explicitly identify the loop nesting depth 
for each BasicBlock.
It assumes a maximum nesting depth of $N$, and replaces the token $BasicBlock$ with 
pre-processed tokens: $BasicBlock\_0$, $BasicBlock\_1$, ..., or $BasicBlock\_N$. 
The input data instances are transformed to use the corresponding token to represent 
basic blocks at different nesting depths.
The data pre-processing introduced in this approach is overly specific to 
the objective function being targeted. 
In a real application scenario 
(e.g.~an optimizing compiler performing many different kinds of analyses),  
this will require transforming or storing a version of the entire input dataset 
for each analysis component, and this can be prohibitively expensive.
Further, this approach is impractical in its assumption 
of a maximum nesting depth.
Even though it is infeasible in practice, 
we include this approach in our experiments 
in order to better understand the learning capabilities of different methods.

\subsubsection{Data Generation}
\label{sec:datagen}

\begin{table}
\small
  \caption{Value ranges in the dataset}
  \label{tab:datagen}
  \begin{tabular}{ccccc}
    \toprule
         & Y      & Sequence & Max loop  & Instructions \\
         & value     &  length & nest level &  in BB\\
    \midrule
    Min & 4,000  & 16 & 1 & 1     \\
    Max & 18,000 & 32 & 5 & 1000      \\
    \bottomrule
  \end{tabular}
  \vspace{-0.1in}
\end{table}
\begin{table}
\small
\caption{Network parameters}
\label{tab:meth}
\begin{tabular}{ccccc}
\toprule
  & LSTM cell  & \# LSTM  & \# FC  & \# FC  \\
  &  size & layers & neurons & layers\\
\midrule
Baseline & 128 & 3 & 256 & 6 \\
Modular & 64 & 1 & 256 & 6 \\
\bottomrule
\end{tabular}
\vspace{-0.1in}
\end{table}

We generate our dataset using the value ranges shown in Table~\ref{tab:datagen}. 
For each labeled example, we randomly choose the sequence length, 
the series of tokens in the sequence, and the number of instructions 
associated with each BasicBlock in the sequence.
We generate 
an equal number of input data instances with maximum loop nesting depth 
equal to 1, 2, 3, 4, and 5.
The total number of instances in the generated dataset are 25,000.
Note that the mapping of inputs to outputs is many-to-one:
the number of instructions (Y-value) depends on 
the program sequence length, 
the size of individual basic blocks, 
the nesting of loops,
and the structural arrangement of basic blocks and loops, 
Also, we allow imperfectly nested loops in program sequences.
These factors allow for a very large number of permutations of 
input data sequences, and make the objective function sufficiently complex 
for evaluating the learning capabilities of different methods.

We want to choose training examples in our dataset 
such that the input data instances 
evenly represent all values 
across the range of the target function being learned.
This is because we want to test the learning capability 
across the entire output value range.
Evenly spreading out the training examples prevents the case 
when learning ends up concentrated in a narrow range of the output space 
as a consequence of the dataset used for training.
To control the impact of the dataset on training, 
we generate the datasets used in our experiments.
Since we treat all output values within a 5\% range of the actual value as correct, 
each actual output value $y$ 
corresponds to a set of correct values with the size of the set proportional to the value $y$.
Therefore, the input data instances are generated using a reciprocal distribution 
which is proportional to $1/y$.
In addition, we experiment with a dataset generated using 
a uniform distribution over values of $y$.

%% file: expmeth.tex
We experimented with the {\em Baseline}, {\em Modular}, 
{\em PartiallyExplicit}, and {\em FullyExplicit} 
methods to compare their learning capability.
Table~\ref{tab:meth} shows network configuration parameters for the different methods
({\em PartiallyExplicit} and {\em FullyExplicit} use the same network structure and size as the 
{\em Baseline} network). 
We varied the training set size in the range 1250 to 21250, 
and determined how the training set size affects accuracy in each case.

We tuned the design of the neural network
for the {\em Baseline} approach 
as much as we could in order to achieve maximum accuracy.
For this purpose, we tried 
different values for multiple parameters 
to learn  their effects on accuracy and found an optimized set of parameter values. 
These parameters include 
batch size, learning rate, optimization algorithm,
dropout rate, 
normalization, data encoding schemes,
network sizes, 
and required number of training steps.
Based on the results of these experiments, 
we fixed the size of the LSTM and FC layers in the {\em Baseline} network
and the {\em Modular} network
as shown in Table~\ref{tab:meth}.

We randomly shuffled the dataset  
and partitioned it into training, test, and validation sets.
We used the validation set to check for accuracy during training, and 
obtained final accuracy numbers using the test set. 
We ran individual experiments multiple times to ensure there are no outliers. 
All code was implemented in Tensorflow version 1.5 on a system equipped with 
NVIDIA Pascal P100 GPUs.

%% file: results.tex
\subsubsection{Comparison of Learning Methods}

\begin{figure}[!ht]
\begin{center}
\includegraphics[width=\linewidth]{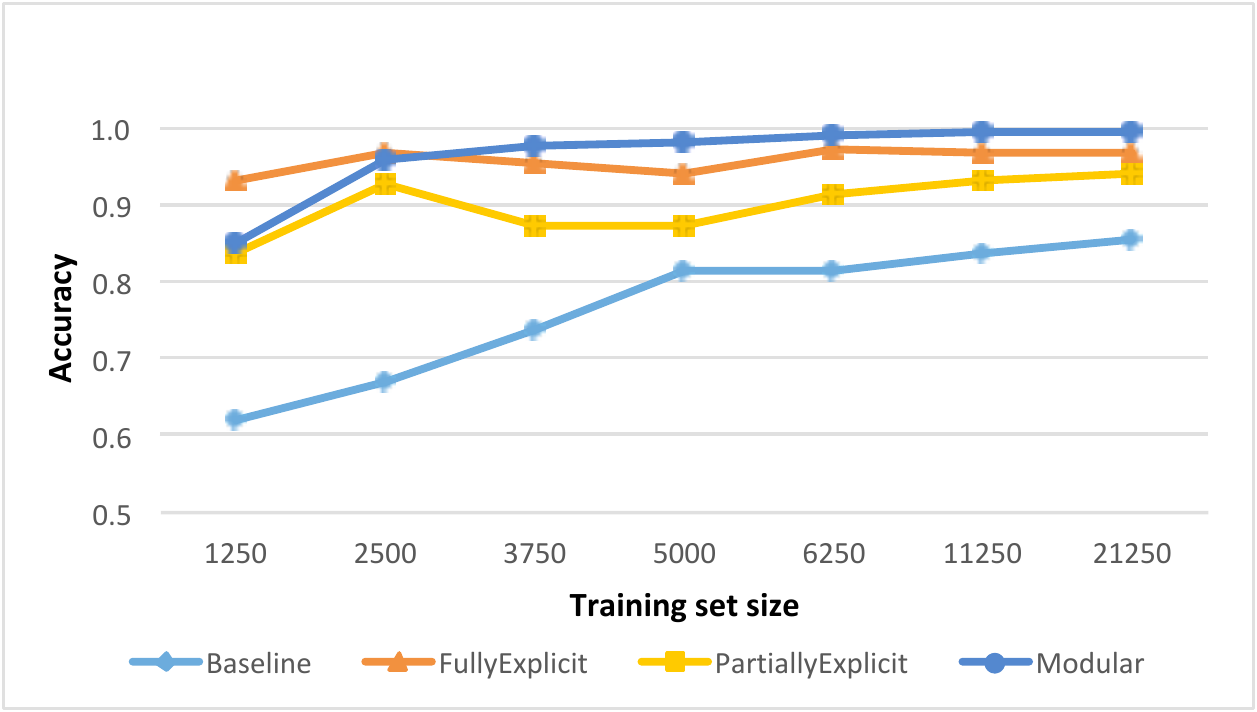}
\caption{Effect of training set size on accuracy}
\label{fig:trainsize}
\end{center}
\vspace{-0.2in}
\end{figure}

We compare the learning capability of the four methods 
described in Section~\ref{sec:lprob}. 
Figure~\ref{fig:trainsize} shows how accuracy varies with 
training set size for each method.
The {\em Modular} network is able to achieve higher accuracy
(up to 29\%) than any of the other approaches.
The {\em FullyExplicit} method achieves accuracy that is close to the {\em Modular} method, 
but it requires pre-processing the input data in a manner 
that is unrealistic in real applications.
Further, the {\em Modular} method achieves its performance with a smaller overall 
network size than the other three methods. 
The {\em PartiallyExplicit} method achieves accuracy that is 
significantly higher than the {\em Baseline} method, indicating that 
providing structural information can simplify the learning process.

\begin{figure*}[!h]
\centering
\subfigure[Accuracy over time for {\em Baseline}]{
\includegraphics[width=.4\textwidth]{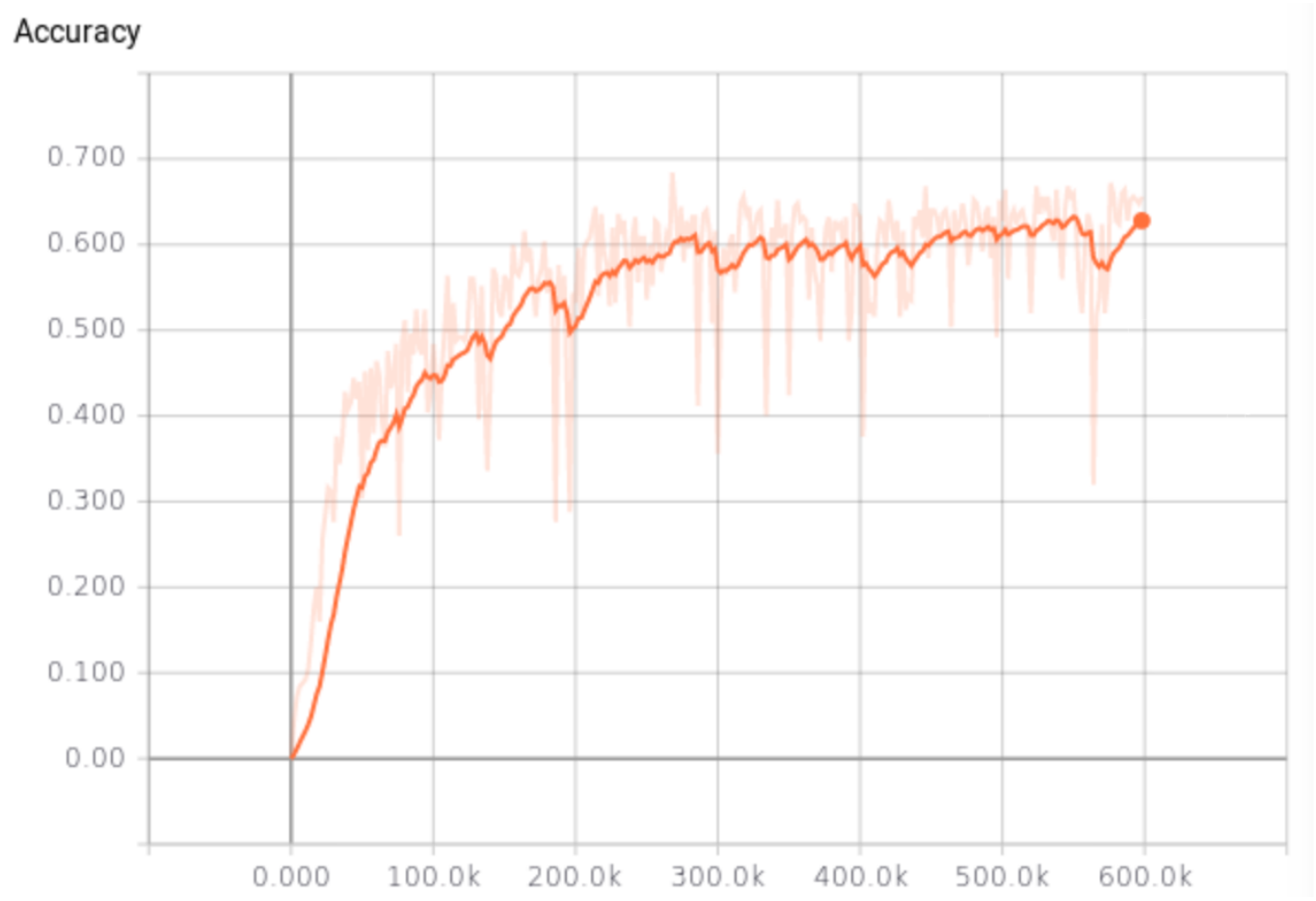}
\label{fig:tradconv}
}
\qquad
\subfigure[Accuracy over time for {\em Modular}]{
\includegraphics[width=.4\textwidth]{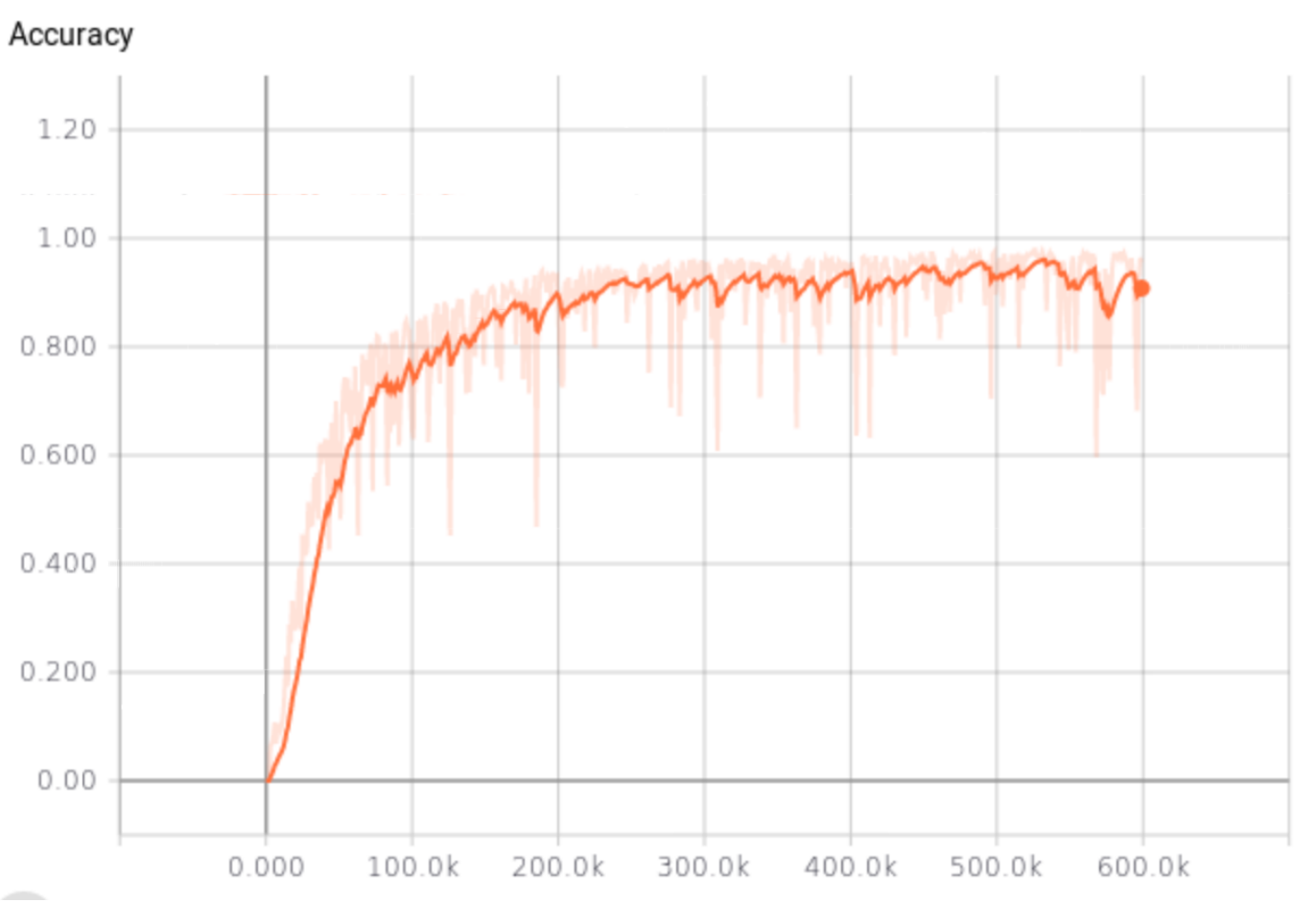}
\label{fig:modconv}
}
\caption{Comparing the convergence of accuracy for {\em Baseline} and {\em Modular} methods}
\label{fig:convergence}
\vspace{-0.1in}
\end{figure*}

We determined that 600,000 iterations were sufficient to 
converge on accuracy for all our training experiments. 
Figure~\ref{fig:tradconv} shows how accuracy varies with training time for the 
{\em Baseline} method  and Figure~\ref{fig:modconv} shows it for 
the {\em Modular} method,  
both for training set size of 2500. 
In this case, the {\em Baseline} method reaches 67\% accuracy 
and the {\em Modular} method reaches 96\% accuracy. 
We observe that both methods converge on accuracy, though the 
{\em Modular} method converges much faster, 
and shows more stable convergence behavior.

\subsubsection{Impact of Input Data Distribution}

\begin{figure*}[!h]
\centering
\subfigure[Reciprocal distribution]{
\includegraphics[width=.42\textwidth]{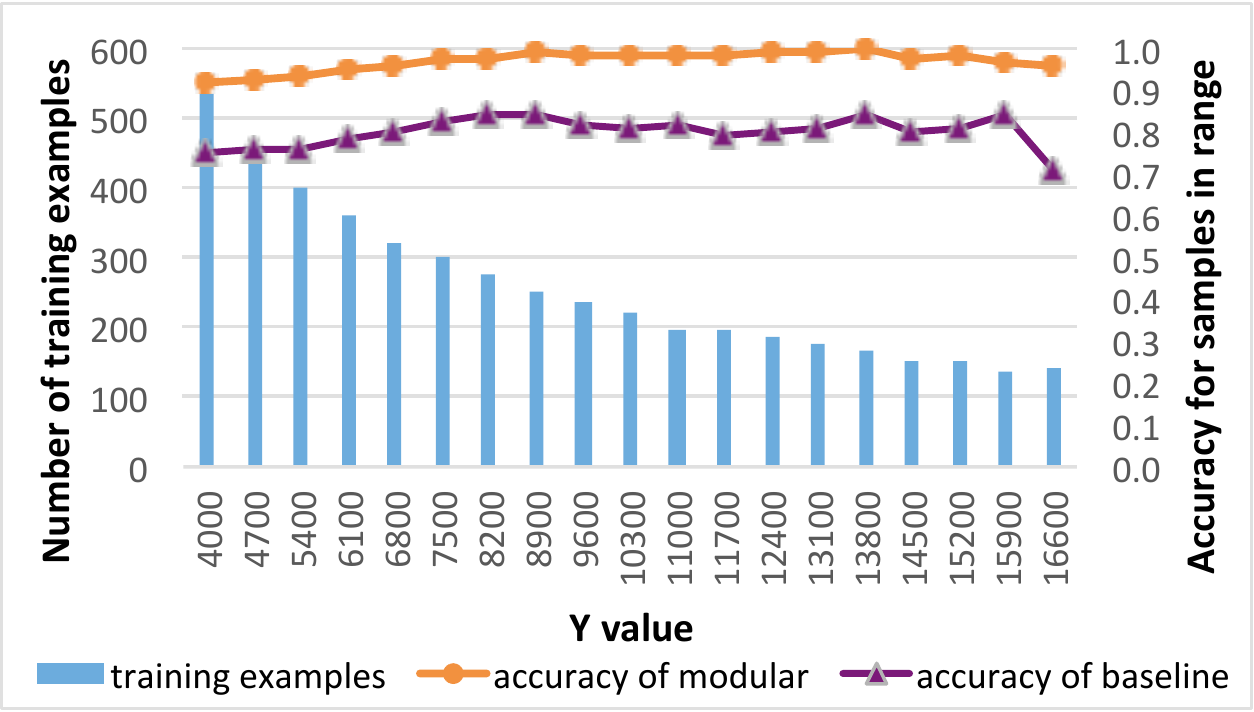}
\label{fig:datadistscal}
}
\qquad
\subfigure[Uniform distribution]{
\includegraphics[width=.42\textwidth]{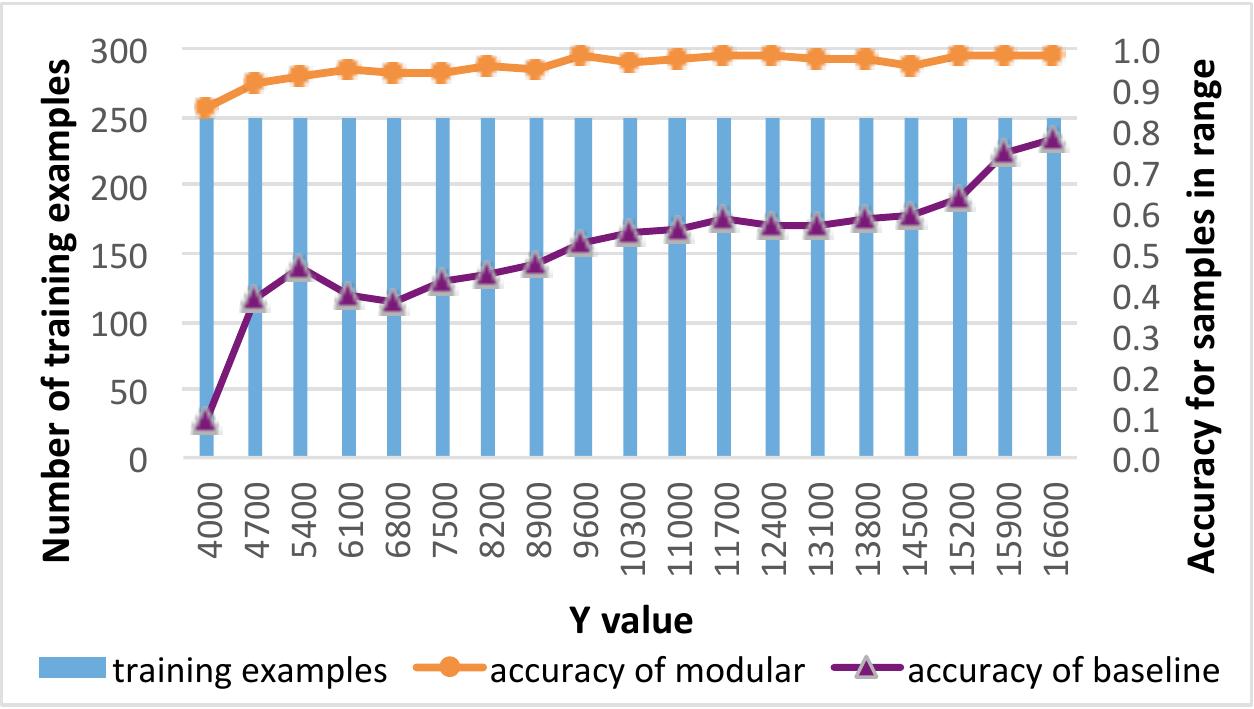}
 \label{fig:datadistunif}
}
\caption{Distribution of training data and corresponding accuracy for output values}
\label{fig:datadist}
\vspace{-0.1in}
\end{figure*}

Figure~\ref{fig:datadistscal} and~\ref{fig:datadistunif} show how the distribution of training examples 
affects accuracy across the range of the function being learned (output Y-values).
The range of Y-values is divided into intervals. 
The plots in the figures are histograms 
showing the number of training examples in each interval, 
together with the accuracy achieved for input data instances 
whose output values are within the interval.
Figure~\ref{fig:datadistscal} 
shows the behavior when using 
the reciprocal distribution as described in Section~\ref{sec:datagen},
with trendlines showing accuracy for both 
the {\em Baseline} and {\em Modular} methods. 
The plot validates that this choice of distribution 
for generating the training data indeed results in uniform learning behavior 
across the target range of the function being learned.
Note that the accuracy is near-constant across the range for both methods.
Figure~\ref{fig:datadistunif}  
shows the behavior when using data generated with 
a uniform distribution across the range of output Y-values.
In this case, the {\em Modular} method still shows near-constant accuracy 
across the range, 
but the accuracy of the {\em Baseline} method varies across the range.
This shows that the {\em Modular} method is more robust 
than the {\em Baseline} method, 
due to its ability to converge faster and achieve higher accuracy with less data.

\subsubsection{Impact of Neural Network Size}
\label{sec:sens}

\begin{figure*}[!h]
\centering
\subfigure[Effect of LSTM size on accuracy]{
\includegraphics[width=.42\textwidth]{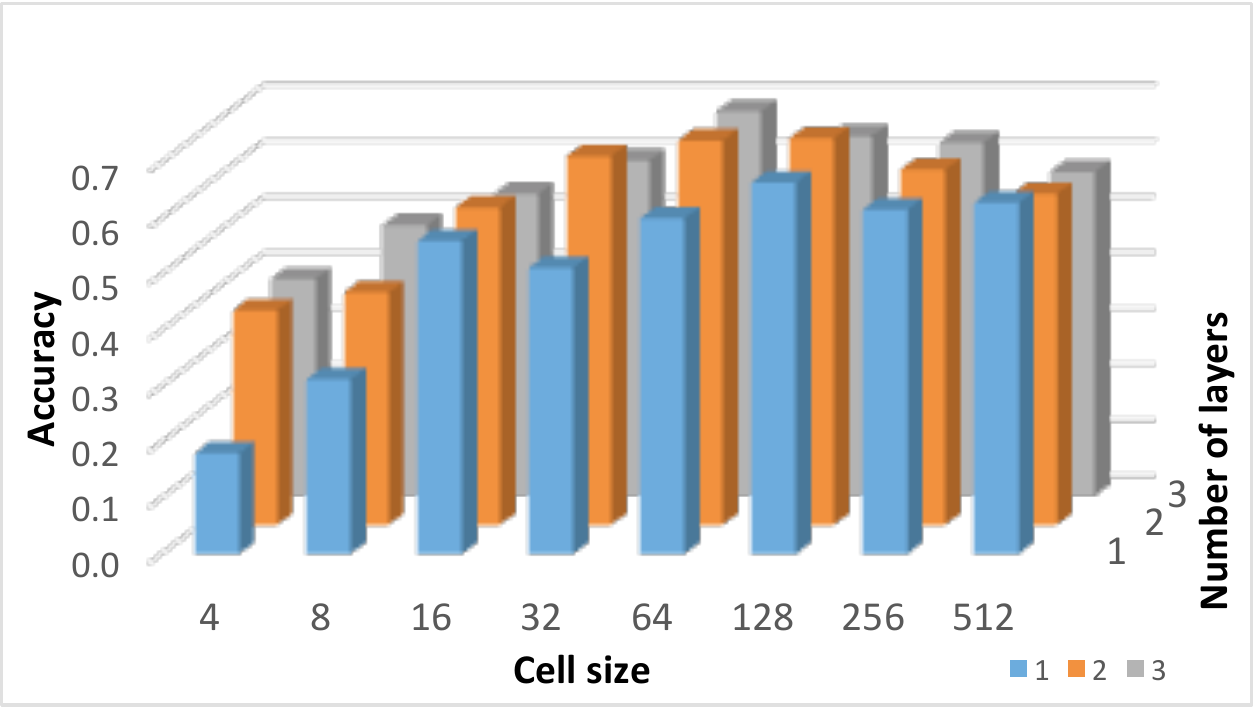}
  \label{fig:rnnsize}
}
\qquad
\subfigure[Effect of FC size on accuracy]{
\includegraphics[width=.42\textwidth]{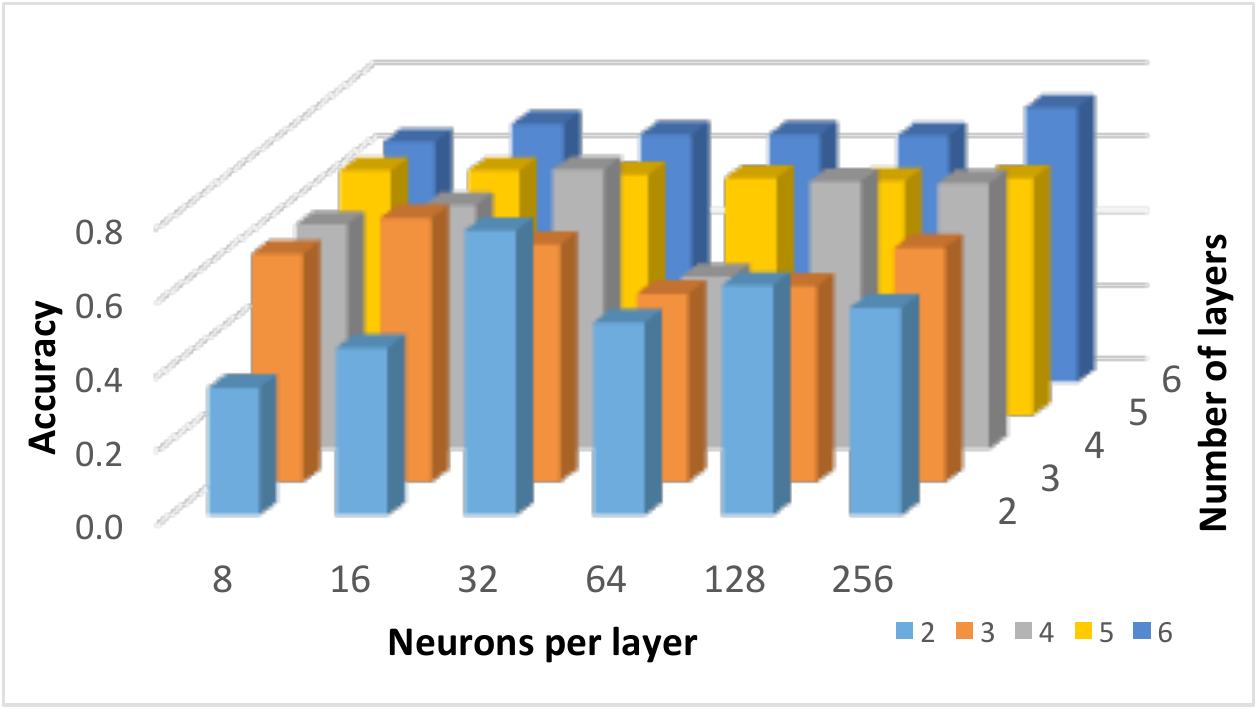}
  \label{fig:fcsize}
}
\vspace{-0.1in}
\caption{Effect of neural network size on accuracy}
\label{fig:nnsize}
\vspace{-0.2in}
\end{figure*}

Figure~\ref{fig:rnnsize} shows how accuracy varies with 
the LSTM cell size and number of LSTM layers for the {\em Baseline} method.
Likewise Figure~\ref{fig:fcsize} shows how accuracy varies with 
the number of neurons per FC layer and the number of FC layers.
We observe that the accuracy does not keep improving beyond a 
certain neural network size. 
On the contrary, the accuracy starts degrading at higher LSTM cell sizes or number of layers, 
as the neural network starts to suffer from overfitting.
Based on trends observed in Figure~\ref{fig:rnnsize} and Figure~\ref{fig:fcsize}, we fixed  
the LSTM cell size, the number of LSTM layers, the number of neurons per FC layer,
and the number of FC layers to the configurations shown in Table~\ref{tab:meth}.
Note that the {\em Modular} network in our experiments uses 
3 LSTM sub-networks, one for each of the 3 identified structures.
Thus, it contains a total of 3 LSTM layers with cell size 64 
and 6 FC layers with 256 neurons per layer.
As a result,
the {\em Modular} network is smaller in size compared to the network used for 
the other three methods.

%% file: related.tex
{\bf Neural Module Networks} \ %
\cite{DBLP:journals/corr/AndreasRDK16} are very similar to our modular neural networks in that they use pre-processing to determine which portions of the input to process through which network component (module) and how to interconnect the outputs and inputs of the various modules. The main difference between this work and ours is that they operate on natural language inputs whereas we operate on computer programs as input. For natural language inputs, the choice of modules and interconnection rules is not clear, and an LSTM is used to learn an approximate mapping. In our case, it is possible to deterministically pre-process the input to map it to a fixed set of semantic structures, which then correspond exactly to the components in our network. 

{\bf Routing Networks} \  %
\cite{rosenbaum2018routing} 
are composed of multiple function blocks and a router that dynamically determines the flow of input through the network. The entire input data is routed to a function block, the result goes back to the router, and the process repeats for a fixed number of iterations. Both the router and the function blocks are trained during learning. Our modular neural network also has a router and multiple component networks, but it works on input sequences, and only partial inputs are forwarded to appropriate component networks. Also, in our case the routing rules are fixed and based on domain knowledge. 

{\bf NLP and Trees}: \  
NLP has a rich body of work exploiting 
syntactic or semantic language structures to better understand 
the contents, intention, and sentiments of texts.
\citet{Mou:2016:CNN:3015812.3016002} proposed
a tree-based CNN to learn explicit structural information in computer programs.
Their model traverses a program AST to extract structural information and then
applies dynamic pooling to gather information over different parts of the tree. 
Their approach to computer programs as structured data resonates with us.
However, 
their model focuses on applying the same CNN to different parts of programs
in a tree-based traversal while our work proposes modularizing the network design
based on structural characteristics.\\
Tree-LSTM~\cite{DBLP:journals/corr/TaiSM15} proposed a tree-structured generalization of the sequential LSTM architecture for improved representation of structured input. The tree-based topology enables capturing structural information in encoded sentence representations and shows better performance than the traditional LSTM on longer sentences. While this work effectively supports our approach of using data representations with structural information, it still relies on LSTM units to learn the structures in input data instead of leveraging them to decompose the networks.   

{\bf Compositional Networks} \  %
\cite{Haykin:1998:NNC:521706, 995117} proposed composing multiple simpler machine learning models 
that address sub-problems to eventually solve a complicated problem.
They also used domain knowledge to design sub-models and control how to
divide input and integrate output.  
However, they partition the training set and build a 
separate specialized model for each set, 
while our work routes structurally different parts within an individual training example
 to specialized models.\\
More recently, \citet{DBLP:journals/corr/abs-1803-03067} proposed a differential architecture that performs structured reasoning by sequencing a new recurrent cell with separate control and memory. The proposed cell performs one reasoning step at its core. Their network solves problems by decomposing them into a sequence of attention-based reasoning operations without relying on externally provided structured representations or specialized modules. 
In contrast, we explicitly try to use known structural information in the design of our neural network.

{\bf Capsule Networks} \  %
\cite{capsulenetworks} have multiple components in the overall network design, and adopt a completely data-driven approach to discovering the structural and contextual information required for learning accuracy. We also use multiple components in our network, but we use domain knowledge to customize the design so that it can more easily capture structural and contextual information.

{\bf Other Approaches}: \  %
\citet{inceptionv4, NIPS2017_7181, DBLP:journals/corr/SerbanSLCPCB16} and \citet{nmt} represent 
recent efforts focused on efficient
construction of networks using structural characteristics of input data
 or learning objectives.
While these approaches share commonality with our work 
in 
building neural networks in a more structured way,
we directly leverage domain knowledge to simplify the task of learning 
structures while they still rely on the  network to automatically 
learn the structure from data. 

{\bf Machine Learning for Program Analysis}: \  %
\citet{2018LitReview} and \citet{2018SurveyPaper} provide a good survey of related work in the area of program optimization.
Some other applications of using machine learning in program analysis include 
clone detection and code similarity analysis~\cite{White:2016:DLClones, Xue:2018:CloneHunter}, 
code security analysis~\cite{Su:2018:Obfuscation}, and 
program synthesis~\cite{nipsEllisRST18, Lee:2018:SearchPgmSynthesis}.
Our work is also in the domain of program analysis, 
but our technique of using structural domain knowledge to improve learning has not been applied before.

%% file: conc.tex
We have presented a method for modular construction of neural networks 
to utilize structure within individual instances of input data, 
and evaluated it on an example code analysis problem.
Compared to the baseline neural network  
(which is based on a tuned LSTM with optimized parameters), 
our modular neural network can achieve higher accuracy using a smaller sized network 
and fewer training examples. 
It also shows more robust learning, performing equally well 
on input data with different distributions, 
whereas the baseline neural network cannot do so.
The method has two main advantages: 
(1) it pre-processes data to expose relevant structural information in the input
which simplifies learning,  
and (2) it modularizes the network structure so it is easier to 
train efficiently and can handle more complex problems.

Even though our work is focused on programming languages and code analysis, 
we note that our method is applicable to other domains with data sources 
that are structured. 
This includes data that is formatted using XML schemas, 
data contained within relational databases, 
and music written using well-known musical notations.